\documentclass{article}
\usepackage{spconf,amsmath,graphicx}

\usepackage{caption} 
\usepackage{algorithm}
\usepackage{algorithmic}
\usepackage{booktabs}
\usepackage{multirow}
\usepackage{amsmath}
\usepackage{amssymb}
\usepackage{mathtools}
\usepackage{amsthm}
\usepackage{stfloats}

\usepackage[caption=false,font=footnotesize,captionskip=0.1mm,farskip=1mm]{subfig}

\usepackage{array,multirow,graphicx}

\newtheorem*{remark}{Remark}

\title{Robustness Against adversarial attacks via Learning Confined Adversarial Polytopes}
\name{Shayan~Mohajer~Hamidi$^*$, Linfeng Ye$^*$\thanks{$^*$ Authors contributed equally.}}
\address{University of Waterloo\\Dept. of Electrical and Computer Engineering\\Waterloo, ON N2L 3G1}

\begin{document}
%
\maketitle
\begin{abstract}
Deep neural networks (DNNs) could be deceived by generating human-imperceptible perturbations of clean samples. Therefore, enhancing the robustness of DNNs
against adversarial attacks is a crucial task. In this paper, we aim to train robust DNNs by limiting the set of outputs reachable via a norm-bounded perturbation added to a clean sample. We refer to this set as adversarial polytope, and each clean sample has a respective adversarial polytope. Indeed, if the respective polytopes for all the samples are compact such that they do not intersect the decision boundaries of the DNN, then the DNN is robust against adversarial samples. Hence, the inner-working of our algorithm is based on learning \textbf{c}onfined \textbf{a}dversarial \textbf{p}olytopes (CAP). By conducting a thorough set of experiments, we demonstrate the effectiveness of CAP
over existing adversarial robustness methods in improving the robustness of models
against state-of-the-art attacks including AutoAttack.
\end{abstract}
\begin{keywords}
Deep Neural Networks, robustness, adversarial training, adversarial attacks, adversarial polytope. 
\end{keywords}
\section{Introduction}
\label{sec:intro}
Recently, generating imperceptible perturbations that can cause a deep neural network (DNN) to make incorrect predictions has become an easy task. These modified samples are referred to as adversarial samples, and the methods used to create such samples are termed adversarial attacks \cite{szegedy2013intriguing}. Adversarial attacks are widely known for their ability to significantly decrease the accuracy of DNNs when tested on adversarial samples \cite{madry2018towards}. When these DNNs are used in critical applications such as autonomous driving, their vulnerability to such adversarial attacks can cause severe security challenges \cite{kurakin2018adversarial,ye2022modeling,ye2021thundernna}. 

A popular method to mitigate the above-mentioned issue, and to increase the robustness of DNNs against adversarial attacks is adversarial training (AT) which can mainly be categorized into two types: (i) to train a classifier by minimizing
the loss for adversarial samples whose simplest yet effective form was first introduced in \cite{madry2018towards} followed by many other modified versions of it \cite{zhang2020attacks,ding2019mma,zhang2020geometry}; and (ii) to minimize the sum of the loss for clean samples and a regularization term for adversarial robustness \cite{zhang2019theoretically,rade2021reducing}. In this paper, we propose an AT method which falls into the second category. Specifically, we devise a regularization term during the training such that it can make the DNNs robust against adversarial attacks. 

Our motivation is based upon observing the DNNs' outputs when they are fed with adversarial samples. In particular, by addition of norm-bounded perturbations to a clean sample, the output of DNN constructs a polytope containing the clean sample, which we refer to as \textit{adversarial polytope} hereafter.

If the \textit{adversarial polytope} for a sample intersects the decision boundary learnt by the DNN, then it is possible to craft an adversarial sample such that the attacked sample is misclassified by the DNN. Therefore, a robust DNN should hopefully learn decision boundaries that do not intersect these polytopes. Yet, if the polytopes for two samples intersect, it is not possible to learn such robust decision boundaries. Therefore, as a remedy, we aim to make the \textit{adversarial polytopes} more compact such that the DNN can learn robust boundaries. To this end, we deploy a regularization term during the training such that it can confine the \textit{adversarial polytopes} for the training samples. To realize this, we take two major steps: (i) we devise a particle-based algorithm to estimate the corner points of the \textit{adversarial polytopes}, and (ii) then, we push these corner points toward the center of the polytope. As such, the polytopes becomes more compact, and therefore, it is possible for the DNN to learn a boundary robust against adversarial attacks. Hence, we train robust DNNs via learning \textbf{c}onfined \textbf{a}dversarial \textbf{p}olytopes (CAP).

By conducting thorough experiments over three different datasets, compared to the state-of-the-art algorithms, we show that the proposed method can yield a high adversarial robustness while maintaining a high clean accuracy.

\noindent $\bullet$ \textbf{Notation}: 
Let $[N]$ denote the set of integers $\{1,2,\cdots,N\}$, and $[N]/n$ is the same set with the element $n$ excluded. In addition, we define $\{x_k\}_{k \in [K]}=\{x_1,x_2,\dots,x_K\}$ for a scalar/vector $x$. Scalars are denoted by lowercase letters (e.g., $x$), and vectors are represented by bold-face lowercase letters (e.g., $\boldsymbol{x}$). Denote by $\boldsymbol{x}[i]$ the $i$-th element of vector $\boldsymbol{x}$. We use $\|\boldsymbol{x}\|_p$ to show the $l_p$ norm of vector $\boldsymbol{x}$. Denote by $\pi_\mathcal{C}(\boldsymbol{x})$ the projection of vector $\boldsymbol{x}$ onto closed set $\mathcal{C}$; that is, $\pi_\mathcal{C}(\boldsymbol{x})=\arg \min_{\boldsymbol{y}}\{\| \boldsymbol{x} -\boldsymbol{y}\|_2 ~|~ y \in \mathcal{C}\}$. Additionally, $\nabla_{\boldsymbol{x}} f(\cdot)$ denotes the gradient vector of function $f(\cdot)$ w.r.t. vector $\boldsymbol{x}$, and $|\mathcal{W}|$ denotes the cardinality of a
set $\mathcal{W}$.

\section{Related work}
Vanilla AT \cite{madry2018towards} uses a simple min-max objective function to train a robust DNN. Specifically, 
the inner maximization generates adversarial examples to maximize the cross-entropy loss, and then model parameters are optimized to minimize the same loss. Although simple and effective, vanilla AT 
decreases accuracy of the DNN for clean samples. To mitigate this issue, TRADES \cite{zhang2019theoretically} devise an objective loss with two terms: (i) the first
term penalizes the cross-entropy loss of clean samples, and (ii) the second term regularizes the
difference between clean output and the output of attacked sample. Later, MART \cite{wang2019improving} assigns more attention to the misclassified samples for robustness. 

AWP \cite{wu2020adversarial} realizes that flatter loss changing with respect to parameter perturbation yields enhanced
generalization of AT, and provides a novel double perturbation mechanism. \cite{carmon2019unlabeled} introduced an approach known as RST, which aims to enhance adversarial training through the utilization of unlabeled data, while also integrating semi-supervised learning technique. \cite{rebuffi2021fixing} focuses on the realm of data augmentation, specifically delving into the effectiveness of employing generative models to enhance performance. Some other papers in the field of adversarial training that focus on (i) model architectures \cite{mao2022towards}, (ii) batch normalization \cite{xie2020adversarial}, and (iii) Distillation from robust models \cite{zi2021revisiting}. 

Our approach to enhancing adversarial robustness, termed as CAP, draws inspiration from recent advances in adversarial training, especially TRADES. We will thoroughly examine and contrast our method with TRADES and other cutting-edge techniques in the field in the Section \ref{sec:exp}.

\section{Methodology} \label{sec:method}
In this section, we elaborate on the inner-workings of CAP. First, we define the meaning of adversarial polytope after discussing some preliminaries.

Let $\boldsymbol{x}_i \in \mathbb{R}^d$ denote the $i$-th training sample in the training set $\mathcal{T}$. The output logits of the DNN (the activations before the softmax function) to $\boldsymbol{x}_i$ is denoted by $\boldsymbol{y}_i \in \mathbb{R}^c$. Therefore, a DNN could be regarded as a function $f_{\boldsymbol{\theta}}: \mathbb{R}^d \rightarrow \mathbb{R}^c$, where $\boldsymbol{\theta}$ is the set of all parameters of the DNN. 

\begin{figure}[t]
	\centering
	\subfloat[]{\includegraphics[width=0.5\columnwidth]{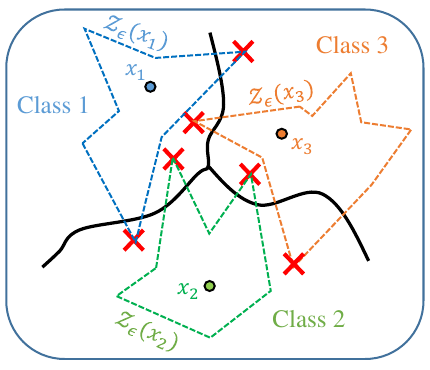}\label{fig1}}
	\subfloat[]{\includegraphics[width=0.5\columnwidth]{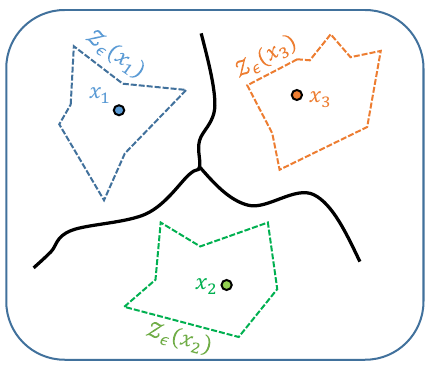}\label{fig2}}
	\vspace{-.15cm}
    \caption{The decision boundary learnt by DNN for three classes. (a) A conventionally trained DNN where the adversarial polytopes cross the decision boundary. (b) A DNN trained by CAP, where the corner points of the polytopes were pushed toward its center making the polytopes more compact.}
	\label{L}
	\vspace{-0.2cm}
\end{figure}

The adversarial polytope for the clean sample $\boldsymbol{x}_i$, denoted by $\mathcal{Z}_{\boldsymbol{\epsilon}(\boldsymbol{x}_i)}$, is the set of all output activations attainable by perturbing $\boldsymbol{x}_i$ by some $\boldsymbol{\epsilon}$ in a predefined constraint $\mathcal{C}$\footnote{The constraint $\mathcal{C}$ could be, for example, an $l_p$ norm constraint.}: 
\begin{align} \label{eq:poly}
\mathcal{Z}_{\boldsymbol{\epsilon}(\boldsymbol{x}_i)}=\{f_\theta (\boldsymbol{x}_i+\boldsymbol{\epsilon}) ~|~ \boldsymbol{\epsilon} \in \mathcal{C} \}   . 
\end{align}
As discussed in Section \ref{sec:intro}, our intention is to make the samples' adversarial polytopes confined such that they do not intersect the decision boundary. To this end, first, we want to figure out which points in $\mathcal{Z}_{\boldsymbol{\epsilon}(\boldsymbol{x}_i)}$ make a DNN vulnerable against adversarial attacks (malignant points). Consider Fig. \ref{fig1}, where the classification boundaries for three classes are depicted by solid black lines. Samples $x_1$, $x_2$, and $x_3$ belong to classes 1, 2, and 3, respectively. The adversarial polytopes for these three samples are depicted using dashed lines. Those areas of adversarial polytopes that cross over the decision boundaries contain malignant points; in particular, an adversary can craft a perturbation to yield a point in these areas, and consequently, causes the DNN to misclassify the underlying sample. 

To tackle this issue, we aim to push the corner points of $\mathcal{Z}_{\boldsymbol{\epsilon}(\boldsymbol{x}_i)}$ toward its center during the training, which in turn pushes the malignant points toward the center of $\mathcal{Z}_{\boldsymbol{\epsilon}(\boldsymbol{x}_i)}$. These corner points are depicted by red crosses in Fig. \ref{fig1}. However, since  $\mathcal{Z}_{\boldsymbol{\epsilon}(\boldsymbol{x}_i)}$ is a non-convex set, finding such corner points is not a straight-forward task. Thus, first, in subsection \ref{sec:detect}, we design an algorithm to find the corner points of $\mathcal{Z}_{\boldsymbol{\epsilon}(\boldsymbol{x}_i)}$.

\begin{algorithm}[tb]
   \caption{Finding the corner points of $\mathcal{Z}_{\boldsymbol{\epsilon}(\boldsymbol{x}_i)}$}
   \label{alg:alternate}
\begin{algorithmic}
   \STATE {\bfseries Input:} Training sample $\boldsymbol{x}_i$, radius $\epsilon$ of the perturbation constraint $\mathcal{C}_{\boldsymbol{\epsilon}}$, step-size $\eta$, number of particles $N$, number of inner optimization steps $T$. 
   \STATE {\bfseries Initialization:} initialize $\{ \boldsymbol{\epsilon}_i^n\}_{i \in N}$ by randomly picking their entries from $U(-\epsilon,\epsilon)$. 
   \STATE Calculate $C_{\boldsymbol{x}_i}$ using \eqref{eq:mean}.
   \FOR{$t=1,2,\dots,T$}
   \FOR{$n=1,2,\dots,N$}
   \STATE $\boldsymbol{\epsilon}_i^n=\boldsymbol{\epsilon}_i^n+\eta \nabla_{\boldsymbol{\epsilon}_i^n} \|f_\theta(\boldsymbol{x}_i+\boldsymbol{\epsilon}_i^n)- C_{\boldsymbol{x}_i}\|_2^2$.
   \STATE $\boldsymbol{\epsilon}_i^n= \pi_{\mathcal{C}_{\boldsymbol{\epsilon}}}(\boldsymbol{\epsilon}_i^n)$.
   \ENDFOR
   \STATE Update $C_{\boldsymbol{x}_i}$ using \eqref{eq:mean}.
   \ENDFOR
   \STATE {\bfseries Output: $\{ \boldsymbol{\epsilon}_i^{*n}\}_{i \in N}$}, $C_{\boldsymbol{x}_i}^*$.
\end{algorithmic}
\end{algorithm}

\subsection{Detecting the corner points of $\mathcal{Z}_{\boldsymbol{\epsilon}(\boldsymbol{x}_i)}$} \label{sec:detect}
In order to identify the corner points of adversarial polytopes, we aim to devise a particle-based method.

Indeed, particle-based methods are widely used in the realm of machine learning. For instance, Stein variational gradient descent (SVGD) \cite{liu2016stein} and the traditional Markov Chain Monte Carlo (MCMC) are two popular particle-based methods used to approximate a target distribution by iteratively updating a set of randomly generated particles.

The crux of our method is to first generate $N$ random particles within each adversarial polytope, and then push these particles to the corner of the polytope. The algorithm is elaborated in details in the following. 

First, we assume a prior distribution $p_{\boldsymbol{\boldsymbol{\epsilon}}}(\cdot)$ for the perturbations. For instance, we assume that the prior distribution for the perturbations is uniform over the perturbation budget; that is, $\boldsymbol{\boldsymbol{\epsilon}}[j] \sim U(-\epsilon, \epsilon)$, where $\epsilon$ is the perturbation budget\footnote{This assumption is also used in other particle-based methods like SVGD.}. Then, for each clean sample $\boldsymbol{x}_i$, $i \in [|\mathcal{T}|]$, we generate $N$ samples from $p_{\boldsymbol{\boldsymbol{\epsilon}}}(\cdot)$, yielding $\{ \boldsymbol{\epsilon}_i^n\}_{n \in [N]}$. Afterward, we add these $N$ perturbations to $\boldsymbol{x}_i$ to generate $N$ perturbed versions of $\boldsymbol{x}_i$. The output of the DNN to these $N$ samples are $\{f_\theta(\boldsymbol{x}_i+\boldsymbol{\epsilon}_i^n)\}_{n \in [N]}$. The empirical center of $\{f_\theta(\boldsymbol{x}_i+\boldsymbol{\epsilon}_i^n)\}_{n \in [N]}$ is calculated as follows:
\begin{align} \label{eq:mean}
C_{\boldsymbol{x}_i}=\frac{1}{N}\sum_{n=1}^N  f_\theta(\boldsymbol{x}_i+\boldsymbol{\epsilon}_i^n).
\end{align}
Then, by iteratively updating $\{ \boldsymbol{\epsilon}_i^n\}_{n \in [N]}$, we aim to push $\{f_\theta(\boldsymbol{x}_i+\boldsymbol{\epsilon}_i^n)\}_{n \in [N]}$ far from $C_{\boldsymbol{x}_i}$ such that they move toward the corner points of $\mathcal{Z}_{\boldsymbol{\epsilon}(\boldsymbol{x}_i)}$. As such, we should maximize the distance between $\{f_\theta(\boldsymbol{x}_i+\boldsymbol{\epsilon}_i^n)\}_{n \in [N]}$ and $C_{\boldsymbol{x}_i}$, where we deploy squared of $l_2$ norm as the underlying distance metric. To maximize the distance, we deploy projected gradient ascent with $T$ iterations and a step-size of $\eta$, and alternatively update each $\{ \boldsymbol{\epsilon}_i^n\}_{n \in [N]}$; that is, to update $\boldsymbol{\epsilon}_i^t$, $t \in [N]$, we fix all the other $\{ \boldsymbol{\epsilon}_i^n\}_{n \in [N]/t}$. After a full iteration over all $\{ \boldsymbol{\epsilon}_i^n\}_{n \in [N]}$, then $C_{\boldsymbol{x}_i}$ would be updated using \eqref{eq:mean}. The procedure for finding the corner points of $\mathcal{Z}_{\boldsymbol{\epsilon}(\boldsymbol{x}_i)}$ is summarized in Algorithm \eqref{alg:alternate}. The outputs of this algorithm are perturbations $\{ \boldsymbol{\epsilon}_i^{*n}\}_{n \in [N]}$ and their corresponding center point $C_{\boldsymbol{x}_i}^*$. Hence, in Algorithm \eqref{alg:alternate}, $\{f_\theta(\boldsymbol{x}_i+\boldsymbol{\epsilon}_i^{*n})\}_{n \in [N]}$ are the representatives for the corner points of $\mathcal{Z}_{\boldsymbol{\epsilon}(\boldsymbol{x}_i)}$.


\subsection{Pushing the corner points toward the center}
Once the corner points of $\mathcal{Z}_{\boldsymbol{\epsilon}(\boldsymbol{x}_i)}$ are estimated by Algorithm \eqref{alg:alternate}, then we shall push those corner points toward its center. To this end, during the training, we minimize the sum of the distances between $\{f_\theta(\boldsymbol{x}_i+\boldsymbol{\epsilon}_i^{*n})\}_{n \in [N]}$ and the empirical center $C_{\boldsymbol{x}_i}^*$. Thus, the loss function in CAP is defined as follows:
\begin{align} \label{eq:loss}
\min_{\theta} \Bigl\{ \text{CE}\left( \text{S} \left(f_\theta(\boldsymbol{x}_i)\right), \boldsymbol{Y} \right) +\lambda \sum_{n=1}^N \|f_\theta(\boldsymbol{x}_i+\boldsymbol{\epsilon}_i^{*n})- C^*_{\boldsymbol{x}_i}\|_2^2\Bigr\},
\end{align}
where $\text{CE}(\cdot,\cdot)$ is the cross-entropy loss, $ \text{S} (\cdot)$ is the softmax function, $\boldsymbol{Y}$ is the label-indicator vector, and $\lambda$ is a regularization parameter. In fact, similarly to TRADES \cite{zhang2019theoretically}, $\lambda$ establishes a trade-off between the clean and robust accuracy.  

\begin{remark}
Although TRADES also uses a regularization term to enforce robustness, however, CAP differs from TRADES in two main aspects: (i) in TRADES, the KL divergence between the output of each clean sample and that of adversarial sample is maximized, yet, CAP maximizes the distance between the center of the adversarial polytope and output of adversarial samples; and (ii) while TRADES only uses one particle, CAP utilizes $N$ particles to better capture the geometry of the adversarial polytopes. 
\end{remark}

\begin{remark}
Considering the inner for-loop over variable $n$ in Algorithm \eqref{alg:alternate}, the optimization over $\boldsymbol{\epsilon}_i^t$, $t \in [N]$, is independent from that over $\boldsymbol{\epsilon}_i^s$, $s\neq t \in [N]$. Therefore, this loop could be performed in parallel. As such, the training time of CAP is controlled by the outer for-loop over $t$, which makes it the same as other AT methods like vanilla AT and TRADES, where they solve a maximization problem using $T$ iterations. 
\end{remark}
\section{Experiments} \label{sec:exp}
In this section, we conclude the paper with several experiments to demonstrate the performance of CAP, and compare its effectiveness with state-of-the-art alternatives under some performance
metrics, including clean (natural) accuracy and adversarial accuracy. 

\begin{table*}[t!]
\begin{center}
\vskip-0.55in
\caption{The clean and robust accuracies of robust ResNet-18 models trained using different adversarial training methods. We use $l_\infty$ attacks with  perturbation budget of $\boldsymbol{\epsilon}=8/255$. Results are obtained based on five runs, where we report the average performance with $95\%$ confidence intervals.}
\vskip -0.1in
\label{tab:main}
 \resizebox{0.8\textwidth}{!}{
 \begin{tabular}{c|ccccccc}
   \toprule
   Dataset & Method  & Clean & FGSM & PGD-20 & PGD-100 & C\&W$_\infty$ & AA  \\
   \midrule
   \multirow{7}{*}{CIFAR-10} & Vanilla AT & 82.78\%          & 56.94\%          & 51.30\%          & 50.88\%          & 49.72\%          & 47.63\%          \\[-0.4em]
    &            & \tiny ~~~$\pm0.12\%$         & \tiny ~~~$\pm0.17\%$  & \tiny ~~~$\pm0.16\%$  & \tiny ~~~$\pm0.26\%$ & \tiny ~~~$\pm0.24\%$ & \tiny ~~~$\pm0.08\%$ \\
    & TRADES & 82.41\%          & 58.47\%          & 52.76\%          & 52.47\%          & 50.43\%          & 49.37\%          \\[-0.4em]
    &            & \tiny ~~~$\pm0.12\%$         & \tiny ~~~$\pm0.19\%$  & \tiny ~~~$\pm0.08\%$  & \tiny ~~~$\pm0.13\%$ & \tiny ~~~$\pm0.17\%$ & \tiny ~~~$\pm0.08\%$ \\
    & MART & 80.70\%          & 58.91\%          & 54.02\%          & 53.58\%          & 49.35\%          & 47.49\%          \\[-0.4em]
    &            & \tiny ~~~$\pm0.17\%$         & \tiny ~~~$\pm0.24\%$  & \tiny ~~~$\pm0.29\%$  & \tiny ~~~$\pm0.30\%$ & \tiny ~~~$\pm0.27\%$ & \tiny ~~~$\pm0.23\%$ \\
    & CAP (ours) & \textbf{83.04\%} & \textbf{59.23\%} & \textbf{54.31\%} & \textbf{54.09\%} & \textbf{50.85\%} & \textbf{50.24\%} \\[-0.4em]
    &            & \tiny ~~~$\pm0.13\%$         & \tiny ~~~$\pm0.25\%$  & \tiny ~~~$\pm0.10\%$  & \tiny ~~~$\pm0.18\%$ & \tiny ~~~$\pm0.12\%$ & \tiny ~~~$\pm0.10\%$
    \\ [-0.2em]
    \midrule
    \multirow{7}{*}{CIFAR-100} & Vanilla AT & 57.27\%          & 31.81\%          & 28.66\%          & 28.49\%          & 26.89\%          & 24.60\%          \\ [-0.4em]
    &            & \tiny ~~~$\pm0.21\%$         & \tiny ~~~$\pm0.11\%$  & \tiny ~~~$\pm0.11\%$  & \tiny ~~~$\pm0.16\%$ & \tiny ~~~$\pm0.08\%$ & \tiny ~~~$\pm0.04\%$ \\ 
    & TRADES & 57.94\%          & 32.37\%          & 29.25\%          & 29.10\%          & 25.88\%          & 24.71\%          \\[-0.4em]
    &            & \tiny ~~~$\pm0.15\%$ & \tiny ~~~$\pm0.18\%$  & \tiny ~~~$\pm0.18\%$  & \tiny ~~~$\pm0.20\%$ & \tiny ~~~$\pm0.16\%$ & \tiny ~~~$\pm0.09\%$ \\
    & MART & 55.03\%          & 33.12\%          & 30.32\%          & 30.20\%          & 26.60\%          & 25.13\%          \\[-0.4em]
    &            & \tiny ~~~$\pm0.10\%$ & \tiny ~~~$\pm0.26\%$  & \tiny ~~~$\pm0.18\%$  & \tiny ~~~$\pm0.17\%$ & \tiny ~~~$\pm0.11\%$ & \tiny ~~~$\pm0.15\%$ \\
    & CAP (ours) & \textbf{58.02\%} & \textbf{33.27\%} & \textbf{30.44\%} & \textbf{30.27\%} & \textbf{26.66\%} & \textbf{25.42\%} \\[-0.4em]
    &            & \tiny ~~~$\pm0.17\%$         & \tiny ~~~$\pm0.15\%$  & \tiny ~~~$\pm0.15\%$  & \tiny ~~~$\pm0.18\%$ & \tiny ~~~$\pm0.15\%$ & \tiny ~~~$\pm0.09\%$ \\ [-0.2em]
    \midrule
    \multirow{7}{*}{SVHN} & Vanilla AT & 89.21\% & 59.81\% & 51.18\% & 50.35\% & 48.39\% & 45.96\% \\[-0.4em]
    &            & \tiny ~~~$\pm0.27\%$         & \tiny ~~~$\pm0.29\%$  & \tiny ~~~$\pm0.29\%$  & \tiny ~~~$\pm0.27\%$ & \tiny ~~~$\pm0.18\%$ & \tiny ~~~$\pm0.21\%$ \\
    & TRADES & 90.20\% & 66.40\% & 54.49\% & 54.18\% & 52.09\% & 49.51\%        \\[-0.4em]
    &            & \tiny ~~~$\pm0.20\%$         & \tiny ~~~$\pm0.18\%$  & \tiny ~~~$\pm0.13\%$  & \tiny ~~~$\pm0.15\%$ & \tiny ~~~$\pm0.10\%$ & \tiny ~~~$\pm0.16\%$ \\
    & MART & 88.70\% & 64.16\% & 54.70\% & 54.13\% & 46.95\% & 44.98\%  \\[-0.4em]
    &            & \tiny ~~~$\pm0.20\%$         & \tiny ~~~$\pm0.24\%$  & \tiny ~~~$\pm0.26\%$  & \tiny ~~~$\pm0.29\%$ & \tiny ~~~$\pm0.24\%$ & \tiny ~~~$\pm0.17\%$ \\
    & CAP (ours) & \textbf{90.26\%} & \textbf{66.63\%} & \textbf{56.01\%} & \textbf{54.50\%} & \textbf{52.39\%} & \textbf{50.19\%} \\[-0.4em]
    &            & \tiny ~~~$\pm0.12\%$         & \tiny ~~~$\pm0.21\%$  & \tiny ~~~$\pm0.14\%$  & \tiny ~~~$\pm0.20\%$ & \tiny ~~~$\pm0.15\%$ & \tiny ~~~$\pm0.12\%$ \\[-0.2em]
   \bottomrule 
 \end{tabular}
 }
\end{center} \vskip -0.22in
\end{table*} 

\noindent $\bullet$ \textbf{Benchmarks}: We compare CAP with vanilla AT \cite{madry2018towards}, TRADES \cite{zhang2019theoretically}, and MART \cite{wang2019improving}. We use $\beta=6$ for TRADES, and $\beta=5$ for MART following their original papers. 

\noindent $\bullet$ \textbf{Datasets}:
In our experiments, we use three popular benchmarks, namely CIFAR-{10,100} \cite{krizhevsky2009learning} and SVHN \cite{netzer2011reading}. 

\noindent $\bullet$ \textbf{Adversarial attacks}: We utilize different white-box attacks including: (i) FGSM \cite{goodfellow2014explaining}, (ii) PGD \cite{madry2018towards} with 20 and 100 iterations, (iii) CW attack \cite{carlini2017towards}, and (iv) AutoAttack (AA) \cite{croce2020reliable}. 

\noindent $\bullet$ \textbf{Training setup}: 
For CIFAR-$\{10,100\}$, we perform adversarial training with a
perturbation budget of $\boldsymbol{\epsilon}= 8/255$ with step-size $\alpha=2/225$, while for SVHN, we set $\alpha=1/225$. We use SGD optimizer with momentum equals 0.9 and a weight decay of 0.0005, and set the batch size equal to 128. We use ResNet-18 as the model. For CIFAR-$\{10,100\}$, we train the model for 120 epochs, with initial learning rate of 0.1 which is divided by 10 at the 80-th and 100-th epochs. On the other hand, for SVHN, we train the model for 80 epochs with initial learning rate of 0.01 that is divided by 10 at the 50-th and 65-th epochs. In addition, we deploy NVIDIA RTX A6000 GPU for training our models. 

\noindent $\bullet$ \textbf{Hyper-parameters for CAP}: We set the regularization parameter $\lambda$ to 0.6, 0.4, and 0.7 for CIFAR-10, CIFAR-100 and SVHN datasets, respectively. Also, we use $N=10$ particles in our experiments. The step-size for projected gradient ascend is $\eta=2/255$, and the number of iterations $T$ equals 40.  

The results are summarized in Table \ref{tab:main}. The results shows that, compared to the benchmark methods, CAP consistently yields higher clean and robust
accuracy over all the datasets. For instance, for CIFAR-10, the other benchmark methods reaches 49.37\% robust accuracy when tested against AA. However, CAP can increase this accuracy to 50.24\%. The same trend is also existing for the other datasets.

\noindent $\bullet$ \textbf{Training on larger models}: 
We further conduct some experiments using large DNNs from Wide-ResNet family \cite{zagoruyko2016wide}; specifically, we trian WRN-34-5 and WRN-34-10 on CIFAR-10 dataset. The training setup, and other hyper-parameters are the same as those used for ResNet-18 model. The results are reported in Table \ref{tab:largemodel}. As seen, CAP outperforms the other benchmarks method in both clean and robust accuracy. This shows that the high robustness provided by CAP is not limited to small models, but could be observed for large models as well. We note that since WRN models yield a high clean accuracy, CAP achieves a slight improvement for this accuracy over the benchmark method.  
\begin{table}[t]
 \caption{Robustness of large models from WRNs family on CIFAR-10 dataset.}
 \vskip -0.2in
 \label{tab:largemodel}
 \begin{center} 
  \resizebox{0.99\linewidth}{!}{
  \begin{tabular}{l|cccc}
    \toprule
    Model & Method & Clean & PGD-20 & AA  \\
    \midrule
    \multirow{2}{*}{WRN-34-5} & TRADES & 83.11\% & 55.78\% & 51.33\% \\
     & CAP (ours) & \textbf{83.13\%} & \textbf{56.15\%} & \textbf{52.07\%} \\
    \midrule
    \multirow{2}{*}{WRN-34-10} & TRADES & 84.80\% & 56.65\% & 52.94\% \\
     & CAP (ours) & \textbf{84.88\%} & \textbf{57.39\%} & \textbf{53.24\%} \\
    \bottomrule
  \end{tabular}
  }
 \end{center} \vskip -0.2in
\end{table}
\section{Conclusion}
Generating imperceptible perturbations in clean samples, deep neural networks (DNNs) can be misled, emphasizing the necessity of making DNNs robust against adversarial attacks. In this paper, we developed an algorithm to train robust DNNs by constraining the range of achievable outcomes through bounded perturbations applied to unperturbed samples, where we referred to these outcomes by adversarial polytope. When this polytope remains compact and avoids intersecting the DNN's decision boundaries, the DNN exhibits resilience against adversarial inputs. As such, our algorithm was based on learning confined adversarial polytopes (CAP). Comprehensive experiments highlight the superiority of CAP compared to existing methods in enhancing model robustness against the cutting-edge attacks.

\bibliographystyle{IEEEbib}
\bibliography{strings,refs}

\end{document}